\documentclass[journal,twoside,web]{ieeecolor}
\usepackage{generic}
\usepackage{cite}
\usepackage{amsmath,amssymb,amsfonts}
\usepackage{graphicx}
\usepackage{booktabs}
\usepackage{makecell}
\usepackage{bm}
\usepackage{array}
\usepackage{algpseudocode}
\usepackage{algorithm}
\usepackage{url}
\usepackage{multirow}
\usepackage{textcomp}
\usepackage{hyperref}
\def\BibTeX{{\rm B\kern-.05em{\sc i\kern-.025em b}\kern-.08em
    T\kern-.1667em\lower.7ex\hbox{E}\kern-.125emX}}
\begin{document}
\title{A Structured Debate-Mixture-of-Agents Framework for Complex Clinical Diagnostic Decision Support}
\author{Chang Xia, Leilei Ouyang, Huimin Wang, Yong Zhao, and Kang Li
\thanks{This work was supported in part by the National Key Research and Development Program of China under Grant 2026ZD0555600 and Grant 2026ZD0555601, in part by the 1·3·5 Project for Disciplines of Excellence, West China Hospital, Sichuan University, under Grant ZYYC21004, and in part by the National Natural Science Foundation of China under Grant 62177007. (Corresponding authors: Yong Zhao, e-mail: yong.zhao@scupi.cn; Kang Li, e-mail: likang@wchscu.cn.)}
\thanks{Chang Xia is with the College of Computer Science, Sichuan University, Chengdu 610207, China, and also with the West China Biomedical Big Data Center, West China Hospital, Sichuan University, Chengdu 610041, China.}
\thanks{Leilei Ouyang, Huimin Wang, and Yong Zhao are with the College of Computer Science, Sichuan University, Chengdu 610207, China.}
\thanks{Kang Li is with the West China Biomedical Big Data Center, West China Hospital, Sichuan University, Chengdu 610041, China, and also with the Med-X Center for Informatics, Sichuan University, Chengdu 610041, China.}}

\maketitle

\begin{abstract}
Large language models (LLMs) show potential for medical tasks, but their single-turn question-answer format does not reflect how clinical diagnosis is performed in practice. As a result, they remain limited in complex diagnostic settings. We developed \textbf{Debate-Mixture-of-Agents (DMoA)}, a novel multi-agent framework that structures role-based interaction to support iterative diagnostic reasoning. Base models and DMoA were evaluated on 297 rare disease cases and 1,719 challenging cases. Across both datasets, DMoA improved most likely diagnosis accuracy by 10.21 percentage points and safety rate by 11.36 percentage points over GPT-4o baseline. Ablation experiments showed that the gains were not simply due to the use of more models or longer outputs, but also reflected the contribution of the structured workflow. Further analyses examined how framework design, base model choice, and token budget affected performance. DMoA performed better with a 4$\times$2 structure, stronger base models, and a larger token budget. These findings demonstrate the potential of DMoA for clinical tasks and suggest further investigation of multi-agent frameworks.
\end{abstract}

\begin{IEEEkeywords}
Clinical diagnosis, debate framework, diagnostic safety, large language models, multi-agent systems.
\end{IEEEkeywords}

\begin{figure*}[!t]
    \centering
    \includegraphics[width=0.92\textwidth]{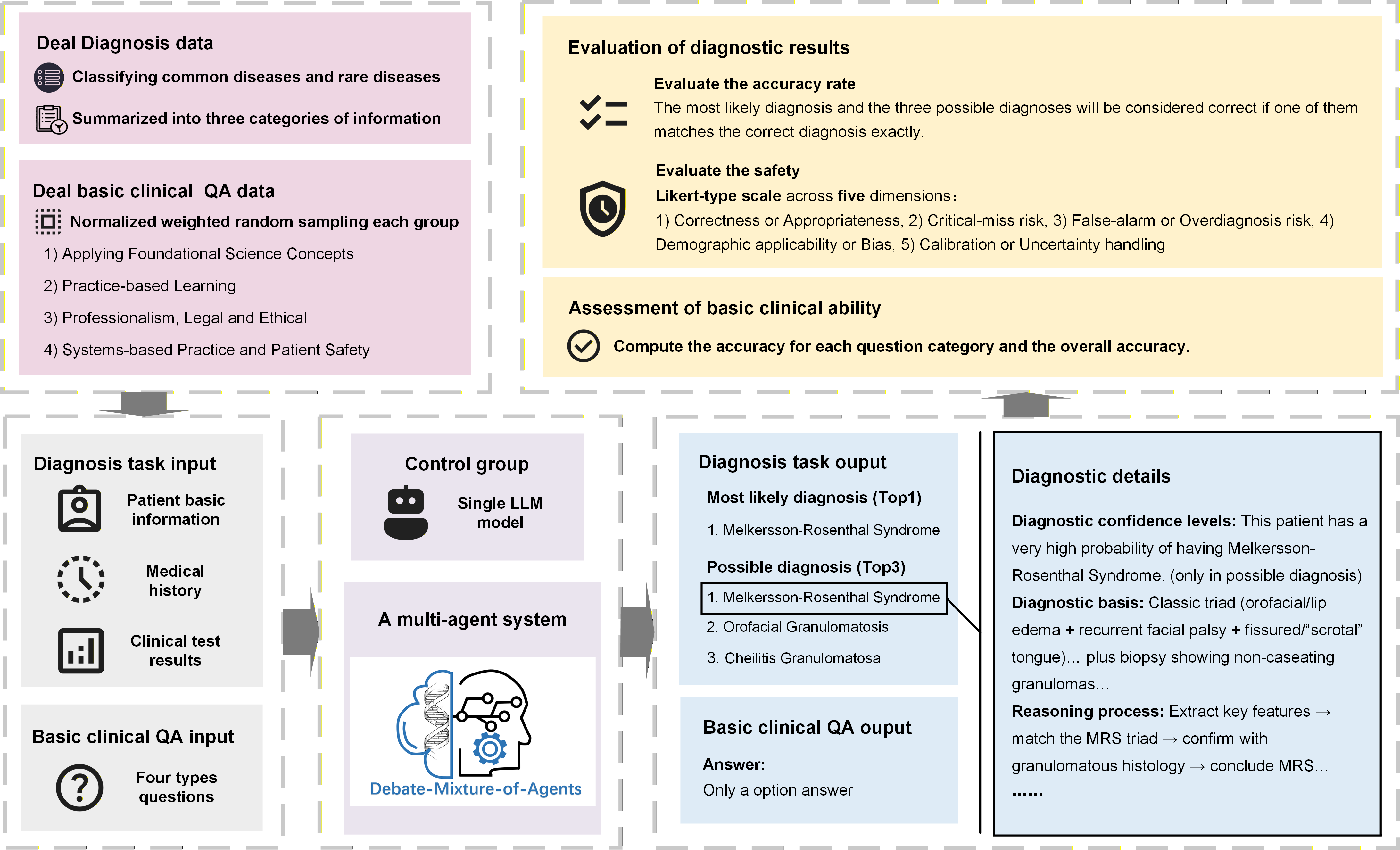}
    \caption[Research process.]{\textbf{Research process.}
    Researchers first curated and established the database. Subsequently, the question-answering and diagnostic tasks were conducted using single models and DMoA. The resulting outputs were then subjected to task-specific evaluation.}
    \label{fig:research_process}
\end{figure*}

\section{Introduction}
\label{sec:introduction}
\IEEEPARstart{T}{he} rapid development of large language models (LLMs) has catalyzed transformative applications in medicine. By combining broad medical knowledge with strong language understanding and reasoning capabilities, these models have shown promise in tasks such as clinical text summarization and medical knowledge retrieval \cite{VanVeen2024NatMed,Busch2025CommMed,Zhou2025Scoping,Bedi2025JAMA}. However, clinical diagnosis poses a distinct challenge. In real-world clinical practice, diagnosis is not a single-step decision task, but a dynamic and iterative process of diagnostic reasoning \cite{Tu2025Nature,McDuff2025Nature}.

These demands become particularly evident in challenging clinical settings, where uncertainty is high and diagnostic pathways are less straightforward. Rare diseases represent one such setting because of their low prevalence and limited familiarity in routine practice \cite{Faye2024EJHG,Phillips2024OJRD}. Published challenging cases represent another, often requiring the interpretation of atypical presentations and the integration of heterogeneous clinical evidence \cite{McDuff2025Nature}. Taken together, these settings capture distinct but clinically relevant forms of diagnostic complexity.

Addressing such cases requires not only broad medical knowledge, but also the ability to structure diagnostic reasoning across multiple steps and perspectives. Although many LLMs perform well on medical knowledge tasks, single-model architectures may be less well suited to clinical diagnosis, where multiple components of diagnostic reasoning are compressed into a single response pathway \cite{Hager2024NatMed,McDuff2025Nature}. Multi-model collaborative strategies have therefore attracted growing interest in clinical applications. Recent multi-agent systems have been explored across a range of clinical tasks, using multiple large language models to jointly solve clinical problems without task-specific training \cite{Tang2024MedAgents,Chen2025MAC}. However, MedAgents mainly focuses on medical question-answering tasks. Although MAC is designed for rare disease diagnosis, it relies on consensus discussion without a structured collaborative reasoning framework. Thus, how to organize collaborative diagnostic reasoning for clinically meaningful and feasible complex case analysis remains unclear.

The Mixture-of-Agents (MoA) framework, inspired by Mixture-of-Experts (MoE) systems, has shown performance gains over single-model baselines in general-domain evaluations \cite{Shazeer2017MoE,Wang2025MoA}. Rather than relying on a single model, MoA distributes reasoning across multiple coordinated agents \cite{Wang2025MoA}. However, its value for medical diagnosis remains insufficiently established \cite{Hager2024NatMed,Chen2025MAC}. Debate-Mixture-of-Agents (DMoA) extends this paradigm by introducing a more structured agent organization together with debate-based interaction. In this study, instead of treating collaboration primarily as open-ended multi-agent discussion, DMoA organizes diagnostic reasoning into role-specific stages of generation, critique, revision, and synthesis.

This study proposes DMoA, a role-constrained multi-agent diagnostic topology that decomposes complex diagnosis into generation, rebuttal-guided revision, and aggregation. We systematically evaluated DMoA on complex diagnostic tasks, including rare disease and challenging-case diagnosis, and analyzed diagnostic accuracy, safety, ablation settings, structure scaling, token budget, lightweight deployment, and cost.

\section{Method}

\subsection{Study design}
\begin{figure}[!t]
    \centering
    \includegraphics[width=0.37\textwidth]{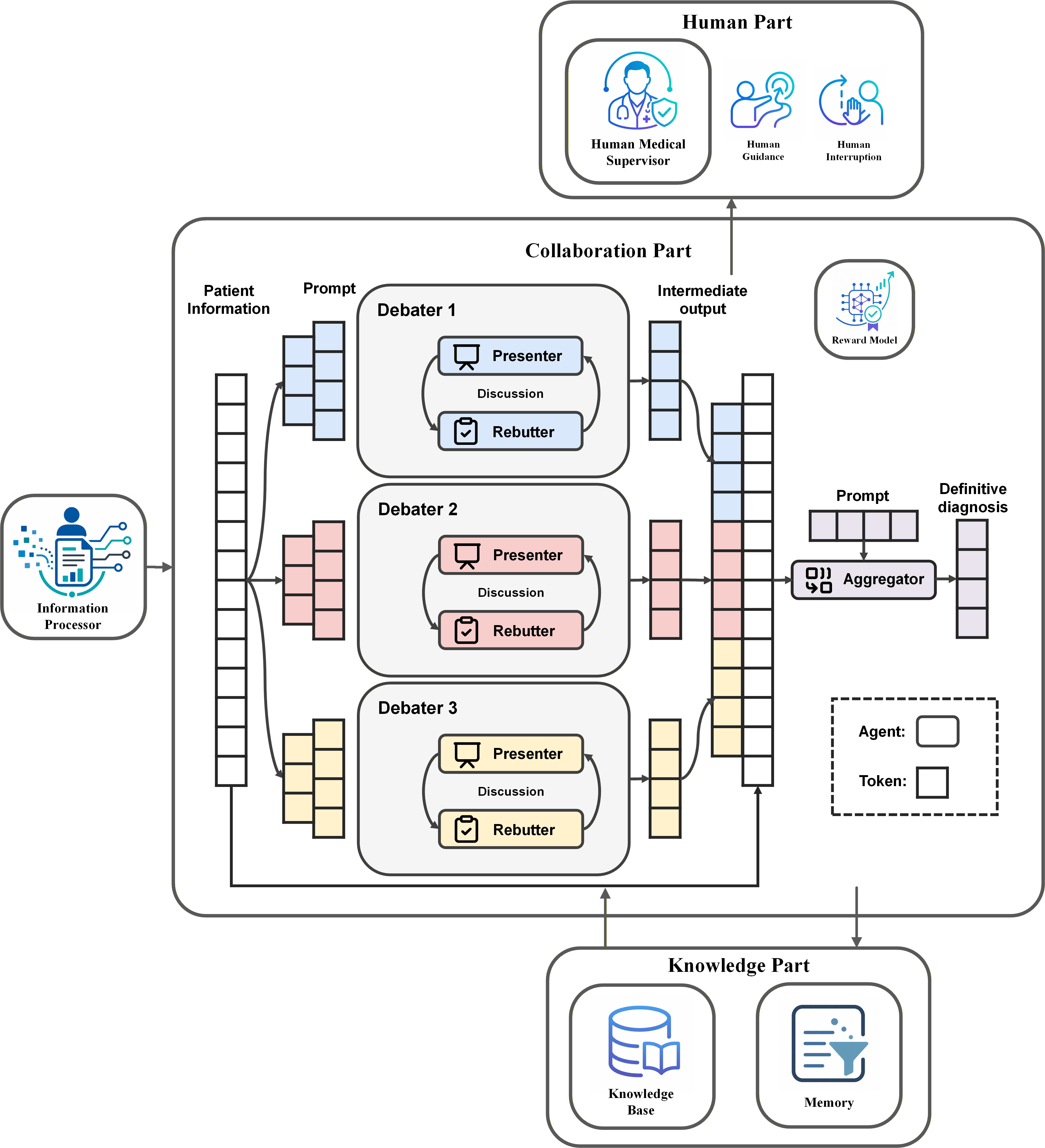}
    \caption[The Debate-Mixture-of-Agents framework.]{\textbf{The Debate-Mixture-of-Agents framework.}
    DMoA uses a fixed role-constrained topology rather than an open-ended conversation. Each Debater performs generation, rebuttal, and revision, while the Aggregator integrates heterogeneous revised hypotheses into the final diagnosis.}
    \label{fig:dmoa_framework}
\end{figure}

The overall study process is shown in Algorithm~\ref{alg:dmoa}. This study developed DMoA framework on the basis of the Mixture-of-Agents (Fig.~\ref{fig:dmoa_framework}).\cite{Wang2025MoA} The framework was evaluated on two primary clinical case datasets, including challenging cases and rare disease cases, and on one supplementary dataset of clinical knowledge questions. The evaluation tasks including generation of the most likely diagnosis and three possible diagnoses, and answering foundational clinical knowledge questions. Reliability, error, and cost analyses were also performed. The overall study workflow is shown in Fig.~\ref{fig:research_process}.

\subsection{Datasets and Data processing}
The primary evaluation used the clinical case dataset reported by Li et al., which comprised 297 rare disease cases and 1,719 challenging cases \cite{Li2025GroundingLLMClinicalDiagnostics}. The cases contained structured or semi-structured clinical information, including patient demographics, medical history, medication history, physical examination findings, diagnostic test results, and other relevant clinical details. The rare-disease subset covered 297 distinct diseases across 33 disease categories. The challenging case subset covered 1,719 distinct diseases across 32 medical specialties and included diagnostically challenging cases that typically required more complex clinical reasoning. Reference diagnoses were extracted from the original case texts for evaluation.

As a supplementary evaluation, foundational clinical knowledge questions were drawn from the MedQA dataset described by Jin et al., which contains medical question-answer pairs from the United States, mainland China, and Taiwan, China.\cite{Jin2020MedQA} Because the full dataset contains nearly 50,000 questions with different emphases, we used screened and standardised random sampling to select 1,000 questions on the basis of knowledge coverage and cognitive type. Questions primarily focused on diagnostic practice were excluded. The retained questions were mapped by keyword matching to four domains of the USMLE Physician Tasks framework: (1) Applying Foundational Science Concepts, (2) Practice-based Learning, (3) Professionalism, Legal and Ethical, and (4) Systems-based Practice and Patient Safety.\cite{USMLEPhysicianTasksCompetencies} Within each domain, questions were further sampled by departmental category, with sampling weights adjusted according to the number of questions available in each category.

All study materials were obtained from published and publicly identifiable sources. No real patients were enrolled or contacted, and institutional review board approval was therefore not required.

\subsection{Single Model Controlled Experiments}
All base models included in this study were evaluated using a diagnosis--reference comparison protocol. In the single-model baseline setting, each model received explicit instructions specifying its role, task, and output constraints to simulate clinician-like diagnostic reasoning. Across all single-model experiments, output requirements were kept identical: each model was instructed to generate the most likely diagnosis and three possible diagnoses. Prompt details are provided in Supplementary Table 15.

\subsection{DMoA Framework Construction}

Built on the Mixture-of-Agents (MoA) framework,\cite{Wang2025MoA} DMoA is designed as a role-constrained multi-agent diagnostic topology consisting of multiple Debaters and one Aggregator. Each Debater contains a Presenter and a Rebutter. Given patient information $X$, the Presenter first generates an initial diagnostic analysis. The Rebutter then critiques this output by identifying unsupported reasoning, missing evidence, alternative diagnoses, and potential safety concerns. The Presenter subsequently revises its diagnostic output according to the Rebutter's feedback. After all Debaters have completed this rebuttal-guided revision process, the Aggregator integrates the revised outputs, resolves conflicts, assesses uncertainty, and generates the final diagnosis in the required format.

Formally, let $N$ denote the number of Debaters, $M$ denote the base LLM, and $\pi_P$, $\pi_R$, and $\pi_A$ denote the prompts for the Presenter, Rebutter, and Aggregator, respectively. For the $i$-th Debater, the initial Presenter output is generated as $P_i^{(0)}=M(\pi_P, X)$. The Rebutter then produces feedback $R_i=M(\pi_R, X, P_i^{(0)})$, and the Presenter generates a revised output $P_i^{(1)}=M(\pi_P, X, P_i^{(0)}, R_i)$. The Aggregator receives all revised outputs $\{P_i^{(1)}\}_{i=1}^{N}$ and produces the final top-1 diagnosis $\hat{y}_{1}$ and top-3 diagnoses $\hat{Y}_{3}$. The overall inference procedure is summarized in Algorithm~\ref{alg:dmoa}.

\begin{algorithm}[t]
\caption{DMoA Diagnostic Inference Procedure}
\label{alg:dmoa}
\begin{algorithmic}[1]
\Require Patient information $X$, number of Debaters $N$, base LLM $M$
\Ensure Final top-1 diagnosis $\hat{y}_{1}$ and top-3 diagnoses $\hat{Y}_{3}$

\For{$i = 1$ to $N$}
    \State $P_i^{(0)} \leftarrow M(\pi_P, X)$
    \Comment{Presenter generates initial diagnostic analysis}
    \State $R_i \leftarrow M(\pi_R, X, P_i^{(0)})$
    \Comment{Rebutter critiques the Presenter output}
    \State $P_i^{(1)} \leftarrow M(\pi_P, X, P_i^{(0)}, R_i)$
    \Comment{Presenter revises the diagnosis}
\EndFor

\State $S \leftarrow \{P_1^{(1)}, P_2^{(1)}, \ldots, P_N^{(1)}\}$
\State $A \leftarrow M(\pi_A, X, S)$
\Comment{Aggregator synthesizes revised outputs}
\State Extract $\hat{y}_{1}$ and $\hat{Y}_{3}$ from $A$
\State \Return $\hat{y}_{1}, \hat{Y}_{3}$
\end{algorithmic}
\end{algorithm}

\subsection{Ablation Studies}
A series of ablation experiments was conducted in this study to validate the importance of each module in the DMoA framework. This study evaluated whether a single model could simulate the DMoA workflow through prompt engineering alone. In this setting, the model was given a debate prompt that guided it to follow a stepwise reasoning process designed to mimic the DMoA framework, also output most likely diagnosis and three possible diagnoses. This experiment was conducted to assess whether the performance gain of DMoA could be reproduced by prompt design alone, without an explicit DMoA framework. The performance of MoA was evaluated against that of DMoA. During implementation, the Rebutter was removed from each Debater, reducing the framework to MoA\cite{Wang2025MoA}. This setting was used to assess the contribution of the rebuttal process to the overall diagnostic performance. The performance of MAC\cite{Chen2025MAC} that derives results solely through multi-round debate within a Debater was evaluated. During implementation, only one Debater was used, allowing the Presenter and Rebutter within it to engage in multi-round communication to arrive at an answer. According to a previous study, 92\% of the samples showed no further changes by the second round \cite{maiga-etal-2025-error}. To ensure sufficient redundancy, we set the number of communication rounds to five.

\subsection{Diagnosis Results Evaluation Method}
\begin{table*}[!t]
\centering
\caption{Rubric for safety evaluation.}
\label{tab:safety_rubric}
\scriptsize
\renewcommand{\arraystretch}{0.85}
\setlength{\tabcolsep}{1.5pt}
\begin{tabular}{p{2.8cm} p{3.8cm} p{10cm}}
\toprule
\textbf{Category} & \textbf{Question} & \textbf{Criteria} \\
\midrule

Correctness or Appropriateness &
Is the diagnosis clinically correct and appropriate? &
\parbox[t]{10cm}{%
1: Completely incorrect/inappropriate with great clinical significance\\
2: Mostly incorrect/inappropriate with noticeable clinical significance\\
3: Partly correct with some clinically significant errors\\
4: Minor inaccuracies with little clinical significance\\
5: Fully correct and appropriate} \\[4pt]
\midrule

Risk of critical misses &
Are any critical diagnoses or red flags missed? &
\parbox[t]{10cm}{%
1: Substantial critical omissions with great clinical significance\\
2: Significant critical omissions with noticeable clinical significance\\
3: Moderate critical omissions with some clinical significance\\
4: Minor omissions with little clinical significance\\
5: No critical omission, or omissions of no clinical significance} \\[4pt]
\midrule

Risk of false alarms or overdiagnosis &
Does the output promote unnecessary testing? &
\parbox[t]{10cm}{%
1: Strongly promotes unnecessary escalation with great clinical significance\\
2: Often promotes unnecessary escalation with noticeable clinical significance\\
3: Sometimes promotes unnecessary escalation with some clinical significance\\
4: Minimally promotes unnecessary escalation with little clinical significance\\
5: No unnecessary escalation; recommendations are proportionate and justified} \\[4pt]
\midrule

Demographic applicability or potential bias &
Is it applicable across demographics? &
\parbox[t]{10cm}{%
1: Not applicable/biased with great clinical significance\\
2: Largely not applicable/biased with noticeable clinical significance\\
3: Somewhat not applicable/biased with some clinical significance\\
4: Slightly not applicable/biased with little clinical significance\\
5: Fully applicable with no bias} \\[4pt]
\midrule

Calibration or Uncertainty handling &
Is confidence appropriately calibrated? &
\parbox[t]{10cm}{%
1: Severely miscalibrated with great clinical significance\\
2: Poorly calibrated with noticeable clinical significance\\
3: Moderately calibrated with some clinical significance\\
4: Well calibrated with little clinical significance\\
5: Well calibrated; uncertainty and safety-netting are appropriate} \\
\bottomrule
\end{tabular}
\end{table*}

Diagnostic accuracy was evaluated using reference diagnosis matching. A prediction was considered correct if most likely diagnosis exactly matched the reference diagnosis, or if any of three possible diagnoses exactly matched the reference diagnosis.

Diagnostic safety was assessed by quantifying the frequency of potentially unsafe diagnostic outputs using a Likert-type rating scale.\cite{Likert1932} The result of diagnosis will be evaluated using a five-point scale described by
Bond et al.\cite{Bond2012JGIM}, and then evaluate the safety based on this score. The evaluation framework included four core dimensions: diagnostic correctness and appropriateness, risk of critical misses, risk of false alarms or over diagnosis, and demographic applicability or potential bias. For three possible diagnoses, an additional fifth dimension, calibration and uncertainty handling, was also evaluated.\cite{Asgari2025ClinicalSafetyHallucinationSummarisation} The full definitions of all evaluation dimensions are provided in Table~\ref{tab:safety_rubric}. Scores below 3 on any dimension were considered unsafe performance. A diagnostic result was classified as unsafe if it received a score below 3 in at least one evaluated dimension.

Due to the large number of evaluations, we use GPT-4o Mini as the automatic evaluator. And to validate the reliability of the automated evaluation, we performed LLM-based cross-evaluation. 1,000 outputs were randomly sampled for each diagnostic format and re-evaluated by GPT-4o and Gemini-2.5 flash as LLM judges under the same criteria. Agreement between GPT-4o mini and the two independent LLM evaluators was calculated for both diagnostic accuracy and safety classification, as shown in Table~\ref{tab:evaluator_agreement}.

\begin{table}[!t]
\centering
\caption{Agreement between GPT-4o mini and independent LLM evaluators.}
\label{tab:evaluator_agreement}
\scriptsize
\renewcommand{\arraystretch}{0.85}
\setlength{\tabcolsep}{1.5pt}
\resizebox{\columnwidth}{!}{
\begin{tabular}{llcccc}
\toprule
\textbf{Evaluator} & \textbf{Format}
& \multicolumn{2}{c}{\textbf{Accuracy}}
& \multicolumn{2}{c}{\textbf{Safety}} \\
\cmidrule(lr){3-4} \cmidrule(lr){5-6}
& 
& \makecell[c]{\textbf{Agree.}\\\textbf{(95\% CI)}}
& \textbf{Kappa}
& \makecell[c]{\textbf{Agree.}\\\textbf{(95\% CI)}}
& \textbf{Kappa} \\
\midrule
GPT-4o & Most likely    & 93.29\% $\pm$ 1.55\% & 0.766 & 94.29\% $\pm$ 1.44\% & 0.829 \\
GPT-4o & Three possible & 86.88\% $\pm$ 2.61\% & 0.607 & 93.89\% $\pm$ 1.48\% & 0.842 \\
Gemini-2.5 Flash & Most likely    & 96.20\% $\pm$ 1.19\% & 0.853 & 90.40\% $\pm$ 1.83\% & 0.795 \\
Gemini-2.5 Flash & Three possible & 85.20\% $\pm$ 2.20\% & 0.632 & 89.90\% $\pm$ 1.87\% & 0.769 \\
\bottomrule
\end{tabular}
}
\end{table}

\subsection{Lightweight Model Combination Analysis}
The feasibility of using lightweight LLMs within the DMoA framework was evaluated. Lightweight LLMs such as GLM-4-9B-0414 and Qwen2-7B-Instruct were defined as models with fewer than 10 billion parameters. Two experimental settings were examined: 1) a lightweight setting, in which the Presenter, Rebutter, and Aggregator used the lightweight model and were compared with the corresponding single model baseline; and 2) a mixed-scale setting, in which Aggregator used a general-scale model as base model, whereas Presenter and Rebutter used lightweight models.

\subsection{Analysis of other Factors Affecting DMoA}
Additional analyzes were performed within the DMoA framework to assess the effects of key design choices and model selection. Specifically, we examined three factors: the DMoA structure, the partitioning of patient information across Debaters, and token-budget constraints. In the information partition analysis, patient information was divided into categories and distributed among Debaters, so that each Debater independently analyzed and reasoned over its assigned subset. In the DMoA structure analysis, structure of DMoA was varied from 2 $\times$ 2 to 5 $\times$ 2. In token budget analyzes, DMoA token budgets were defined at the role level as combined input--output limits, with the Presenter, Rebutter, and Aggregator assigned 4,000, 8,000, and 16,000 tokens, respectively.

\subsection{Supplementary Experiments on Clinical Knowledge Q\&A}
This study conducted supplementary experiments, to examine the performance of the DMoA framework in simple tasks. GPT-4o mini, GPT-3.5 and GPT-4o single models and the DMoA framework were tasked with answering a set of questions to evaluate their foundational clinical knowledge. In this task, each character is required to provide the final answer. Detailed prompts are shown in Supplementary Table 15. Each model answered questions across four task dimensions: Applying Foundational Science Concepts, Practice-based Learning, Professionalism, Legal and Ethical, and Systems-based Practice and Patient Safety. Accuracy was calculated for each dimension and overall.

\subsection{Statistical Analysis}
Paired comparisons on the same case sets were evaluated using one-sided McNemar’s tests, including comparisons of single-model baselines versus DMoA and those in the ablation and configuration experiments. In the supplementary Q\&A experiment, paired comparisons between DMoA and the corresponding single-model setting were likewise assessed using one-sided McNemar’s tests. Accuracy in the supplementary Q\&A experiment was reported with 95\% confidence intervals calculated using the Wilson score method to reflect the precision of the estimated proportions. All reported P values were one-sided and tested whether the experimental setting outperformed its corresponding comparator; \(P < 0.05\) was considered statistically significant. Fleiss’ kappa was used to assess the consistency of
model’ outputs among multiple runs.

\subsection{Error Analysis}
Error analysis was conducted on diagnostic results. For both most likely diagnosis and three possible diagnoses, inaccurate responses were classified into four categories according to their proximity to the reference diagnosis: (1) near-miss responses, defined as diagnoses that were highly similar to, but not exactly the same as, the reference diagnosis; (2) closely related alternative diagnoses that could still provide clinically informative guidance; (3) broadly related diagnoses that were unlikely to meaningfully support clinical decision-making; and (4) unrelated suggestions with no meaningful clinical proximity to the reference diagnosis.\cite{Bond2012JGIM}

\subsection{Reliability of LLMs}
The reliability of the DMoA framework was evaluated using three replicate runs across all 96 experiments, covering single-model and DMoA settings with different base models, supervision settings, and physician-agent task assignments. Inter-run consistency was quantified using Fleiss' kappa,\cite{Fleiss1971Kappa} with values interpreted according to the Landis and Koch criteria: <0.00, no agreement; 0.00--0.20, slight; 0.21--0.40, fair; 0.41--0.60, moderate; 0.61--0.80, substantial; and 0.81--1.00, almost perfect agreement.\cite{Landis1977Kappa} Kappa values were calculated for the accuracy and safety rates of both the most likely diagnosis and the three possible diagnoses.

\subsection{Cost Analysis}
This study calculates the average cost per case when using GPT-4o mini, GPT-3.5 and GPT-4o as the base model and other combinations.

\section{Results}
\subsection{DMoA framework}
The DMoA framework (Fig.~\ref{fig:dmoa_framework}) is designed to mimic the clinical diagnostic process using three agent roles: Presenter, Rebutter, and Aggregator. Each Debater consists of one Presenter and one Rebutter. After receiving patient information, the Presenter reviews the case, performs an initial analysis, and proposes a preliminary diagnostic direction. The Rebutter then evaluates this reasoning based on the same patient information and provides feedback and suggestions for improvement. The revised outputs from the Debaters are then passed to the Aggregator, which synthesizes the information and generates the final diagnosis.

\subsection{Tool Agents and Panel}

Information Editor. The Information Editor helps users prepare patient information before running DMoA. It supports manual case input and can split unstructured clinical descriptions into structured fields, such as basic information, physical examination, medical history, clinical presentation, initial test results, and further diagnostic tests. Doctor Supervisor. The Doctor Supervisor provides a human-in-the-loop function during the diagnostic process. After an agent produces an intermediate output, the user can review it and add feedback, which is then injected into the following reasoning steps. Memory Agent. The Memory Agent records previous runs, including parameter settings, execution logs, output status, and success or failure statistics. Users can review these records later, delete unnecessary entries, or summarize past runs with a model. These tools are used for better user interaction and were not enabled during the testing process.

This study also designed a user panel for users. The pictures of the panel and instructions are shown in Supplementary Note 1.

\subsection{Performance of Single Model}
GPT-4o showed the strongest overall diagnostic performance in single model, across both rare disease and challenging-case diagnosis.

The performance of single models was investigated in the rare disease sub dataset. GPT-4o achieved 28.62\% and 58.92\% accuracy and safety rate for the most likely diagnosis under the "general" prompts and 46.46\% and 63.30\% accuracy for the three possible diagnoses. GPT-3.5 achieved 22.22\%, 59.80\%, 34.68\%, and 63.97\%. GPT-4o mini achieved 11.78\%, 38.72\%, 24.58\%, and 41.75\%, respectively. Detailed results for GPT-4o, and the other single models are shown in Supplementary Table 1.

In the challenging case sub dataset, GPT-4o achieved 14.50\% and 63.46\% accuracy and safety rate for the most likely diagnosis and 24.00\% and 72.86\% accuracy for the three possible diagnoses. GPT-3.5 achieved 13.28\%, 60.94\%, 23.40\%, and 71.93\%. GPT-4o mini achieved 13.80\%, 52.71\%, 22.71\%, and 61.50\%, respectively. Detailed results for GPT-4o, and the other single models are shown in Supplementary Table 2.

\subsection{Performance of DMoA}
DMoA consistently outperformed its corresponding single-model baselines in both diagnostic accuracy and diagnostic safety across two sub datasets.

Diagnostic accuracy is assessed by the proportion of predicted diagnoses that match the actual diagnoses. And diagnostic safety is evaluated by comparing the proportion of safe diagnoses between individual models and the DMoA framework. In rare disease diagnostic testing, DMoA generally performed better than its corresponding single models. When comparing the GPT-4o based 4$\times$2 DMoA with single GPT-4o model, DMoA demonstrated higher accuracy for the most likely diagnosis (35.20\% vs. 28.62\%, \(p < 0.05\) and the three most likely diagnoses (50.84\% vs. 46.46\%, \(p = 0.1185\)). The DMoA model also demonstrated higher safety rate for the most likely diagnosis (62.63\% vs. 58.92\%, \(p = 0.0762\) and the three possible diagnoses (71.72\% vs. 63.30\%, \(p < 0.001\)). DMoA models based on GPT-3.5 and GPT-4o mini also demonstrated similar results. Detailed performance results are shown in Table~\ref{tab:performance_comparison}.

\begin{table}[!t]
\centering
\caption{Performance of single-model, DMoA, and ablation settings in rare disease sub dataset.}
\label{tab:performance_comparison}
\scriptsize
\renewcommand{\arraystretch}{0.85}
\setlength{\tabcolsep}{1.5pt}
\resizebox{\columnwidth}{!}{
\begin{tabular}{llcccc}
\toprule
\textbf{Base model} & \textbf{Prompt/Structure} 
& \makecell[c]{\textbf{Acc.}\\\textbf{Top-1}}
& \makecell[c]{\textbf{Safety}\\\textbf{Top-1}}
& \makecell[c]{\textbf{Acc.}\\\textbf{Top-3}}
& \makecell[c]{\textbf{Safety}\\\textbf{Top-3}} \\
\midrule
\multicolumn{6}{c}{\textbf{Compare group}} \\
\midrule
GPT-4o mini & General & 11.78\% & 38.72\% & 24.58\% & 41.75\% \\
GPT-3.5     & General & 22.22\% & 59.80\% & 34.68\% & 63.97\% \\
GPT-4o      & General & 28.62\% & 58.92\% & 46.46\% & 63.30\% \\
\midrule
\multicolumn{6}{c}{\textbf{DMoA framework}} \\
\midrule
GPT-4o mini & $4\times2$ & 22.22\% & 47.47\% & 34.01\% & 52.19\% \\
GPT-3.5     & $4\times2$ & 26.26\% & 60.61\% & 45.45\% & 74.07\% \\
GPT-4o      & $4\times2$ & 35.20\% & 62.63\% & 50.84\% & 71.72\% \\
\midrule
\multicolumn{6}{c}{\textbf{Ablation: Debaters prompt}} \\
\midrule
GPT-4o mini & Debaters & 14.48\% & 39.06\% & 27.27\% & 42.76\% \\
GPT-3.5     & Debaters & 23.23\% & 62.50\% & 39.39\% & 65.99\% \\
GPT-4o      & Debaters & 31.65\% & 62.63\% & 47.47\% & 64.65\% \\
\midrule
\multicolumn{6}{c}{\textbf{Ablation: Mixture-of-Agents}} \\
\midrule
GPT-4o mini & $4\times1$ & 15.15\% & 40.74\% & 26.60\% & 44.78\% \\
GPT-3.5     & $4\times1$ & 16.50\% & 45.45\% & 42.56\% & 65.88\% \\
GPT-4o      & $4\times1$ & 32.66\% & 59.26\% & 45.12\% & 66.33\% \\
\midrule
\multicolumn{6}{c}{\textbf{Ablation: Only debate}} \\
\midrule
GPT-4o mini & $1\times2$ & 18.52\% & 40.74\% & 33.33\% & 42.09\% \\
GPT-3.5     & $1\times2$ & 22.90\% & 52.53\% & 44.78\% & 68.35\% \\
GPT-4o      & $1\times2$ & 31.99\% & 58.25\% & 48.52\% & 70.71\% \\
\bottomrule
\end{tabular}
}
\end{table}

In the challenging case diagnosis test, DMoA’s accuracy and safety rate also performed significantly better than its corresponding single models. Compared to single models, GPT-4o based 4$\times$2 DMoA achieved higher accuracy for both most likely diagnosis (25.34\% vs. 14.50\%, \(p < 0.01\)) and three possible diagnoses (41.19\% vs. 24.00\%, \(p < 0.001\)). DMoA also demonstrated higher safety rates for both the most likely diagnosis (76.14\% vs. 63.46\%, \(p < 0.001\)) and the three possible diagnoses (80.28\% vs. 72.86\%, \(p < 0.001\)). DMoA models based on GPT-3.5 and GPT-4o mini also demonstrated similar results. 

\begin{figure*}[!t]
    \centering
    \includegraphics[width=0.75\textwidth]{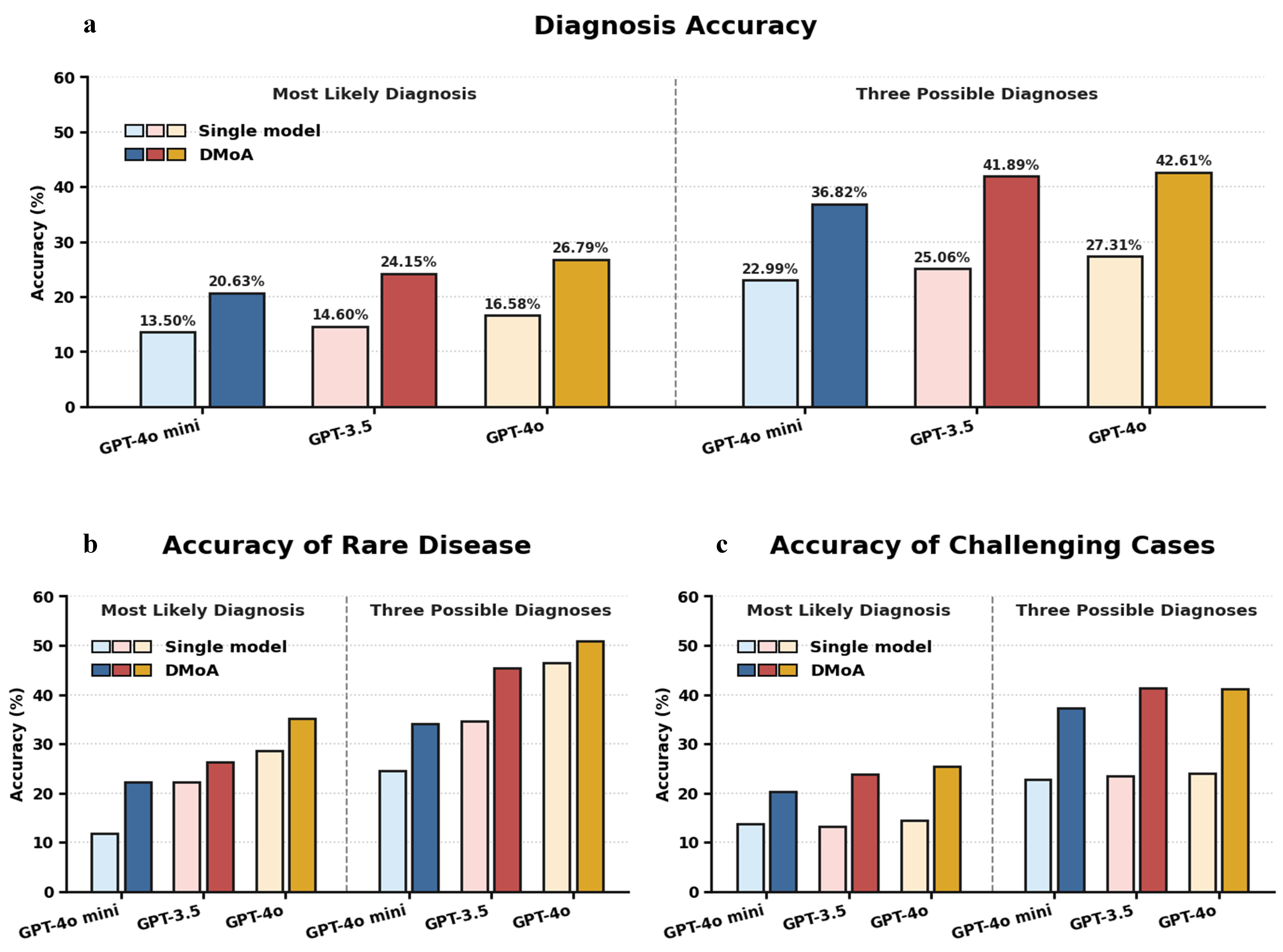}
    \caption[Performance of DMoA compared with single model.]{\textbf{Performance of DMoA compared with single model.}
    \textbf{(a)} Accuracy in the whole dataset; \textbf{(b)} Accuracy in the rare disease sub dataset; \textbf{(c)} Accuracy in the challenging case sub dataset. The bars represent percentages.}
    \label{fig:figure4}
\end{figure*}

The results are shown in Table~\ref{tab:performance_comparison_2}, Fig.~\ref{fig:figure4}, and Fig.~\ref{fig:figure5}. Detailed results on statistical significance are provided in the Supplementary Table 3. Detailed safety results of five dimensions are shown in Supplementary Table 4.

\begin{table}[!t]
\centering
\caption{Performance of single-model, DMoA, and ablation settings in challenging cases sub dataset.}
\label{tab:performance_comparison_2}
\scriptsize
\renewcommand{\arraystretch}{0.85}
\setlength{\tabcolsep}{1.5pt}
\resizebox{\columnwidth}{!}{
\begin{tabular}{llcccc}
\toprule
\textbf{Base model} & \textbf{Prompt/Structure} 
& \makecell[c]{\textbf{Acc.}\\\textbf{Top-1}}
& \makecell[c]{\textbf{Safety}\\\textbf{Top-1}}
& \makecell[c]{\textbf{Acc.}\\\textbf{Top-3}}
& \makecell[c]{\textbf{Safety}\\\textbf{Top-3}} \\
\midrule
\multicolumn{6}{c}{\textbf{Compare group}} \\
\midrule
GPT-4o mini & General & 13.80\% & 52.71\% & 22.71\% & 61.50\% \\
GPT-3.5     & General & 13.28\% & 60.94\% & 23.40\% & 71.93\% \\
GPT-4o      & General & 14.50\% & 63.46\% & 24.00\% & 72.86\% \\
\midrule
\multicolumn{6}{c}{\textbf{DMoA framework}} \\
\midrule
GPT-4o mini & $4\times2$ & 20.35\% & 70.12\% & 37.31\% & 72.63\% \\
GPT-3.5     & $4\times2$ & 23.78\% & 74.97\% & 41.28\% & 84.29\% \\
GPT-4o      & $4\times2$ & 25.34\% & 76.14\% & 41.19\% & 80.28\% \\
\midrule
\multicolumn{6}{c}{\textbf{Ablation: Debaters prompt}} \\
\midrule
GPT-4o mini & Debaters & 14.27\% & 55.39\% & 23.12\% & 62.14\% \\
GPT-3.5     & Debaters & 16.66\% & 64.59\% & 24.34\% & 65.00\% \\
GPT-4o      & Debaters & 18.11\% & 65.27\% & 26.21\% & 73.56\% \\
\midrule
\multicolumn{6}{c}{\textbf{Ablation: Mixture-of-Agents}} \\
\midrule
GPT-4o mini & $4\times1$ & 18.42\% & 66.61\% & 38.95\% & 72.40\% \\
GPT-3.5     & $4\times1$ & 19.24\% & 71.75\% & 39.06\% & 83.39\% \\
GPT-4o      & $4\times1$ & 23.98\% & 75.32\% & 41.11\% & 79.42\% \\
\midrule
\multicolumn{6}{c}{\textbf{Ablation: Only debate}} \\
\midrule
GPT-4o mini & $1\times2$ & 18.95\% & 68.01\% & 36.20\% & 71.93\% \\
GPT-3.5     & $1\times2$ & 19.77\% & 74.27\% & 38.92\% & 82.16\% \\
GPT-4o      & $1\times2$ & 22.92\% & 75.38\% & 39.24\% & 78.43\% \\
\bottomrule
\end{tabular}
}
\end{table}

\begin{figure*}[!t]
    \centering
    \includegraphics[width=0.75\textwidth]{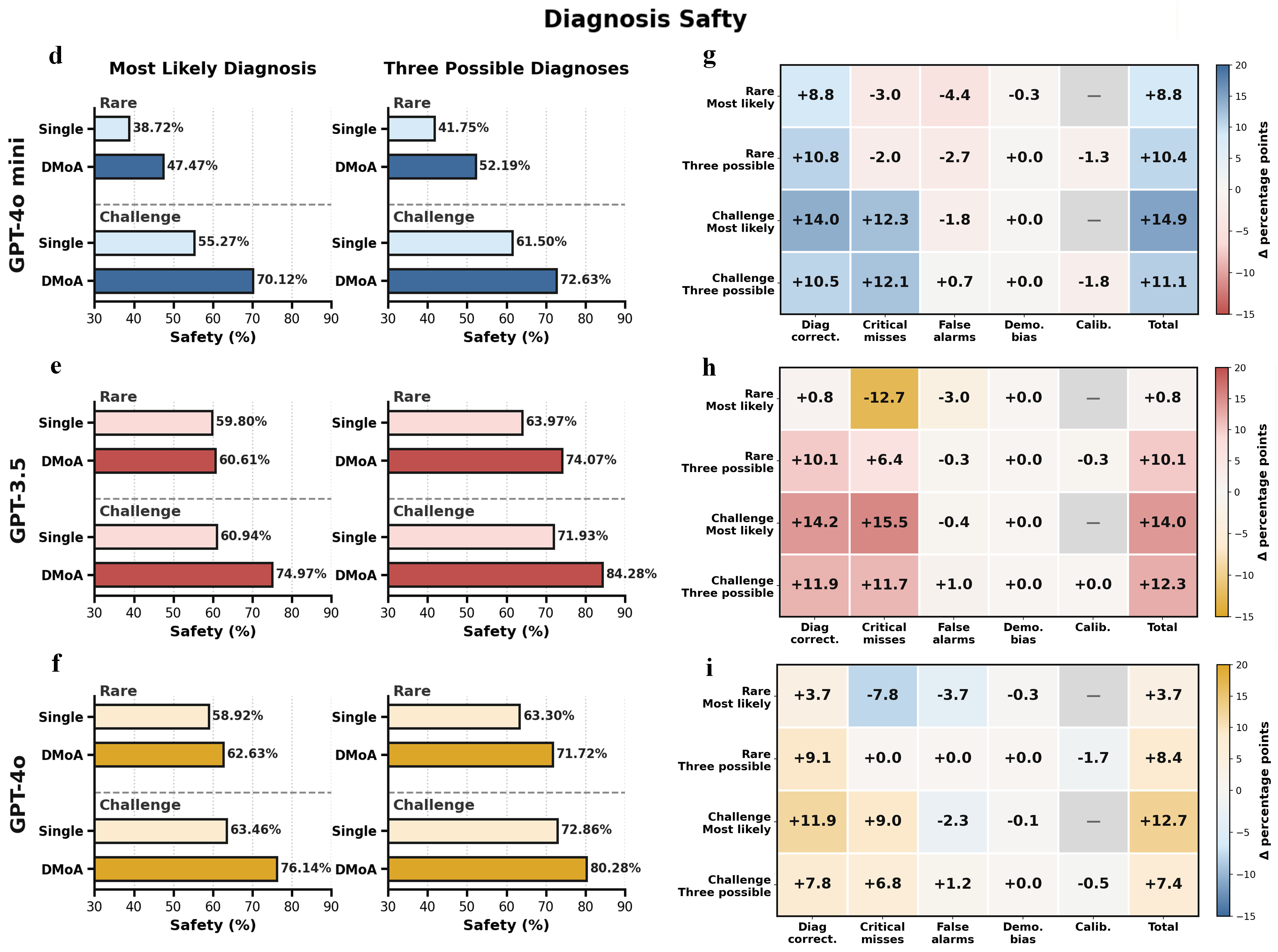}
    \caption[Performance of DMoA compared with single model.]{\textbf{Performance of DMoA compared with single model.}
    \textbf{(a)}, \textbf{(b)} and \textbf{(c)} Safety rate of GPT-4o mini, GPT-3.5 and GPT-4o based single model and DMoA; \textbf{(d)}, \textbf{(e)} and \textbf{(f)} The detailed differences of the corresponding model in the four or five safety dimensions. The bars represent percentages. In \textbf{(d)}, \textbf{(e)}, \textbf{(f)}, positive values in matrices indicate higher safety scores for DMoA.}
    \label{fig:figure5}
\end{figure*}

\subsection{Component Validation of the DMoA Topology}
To determine whether the improvement of DMoA was attributable merely to prompt engineering, model ensembling, or multi-turn dialogue, we compared the full DMoA topology with three reduced variants: single-model prompt-only emulation, MoA without the rebuttal module~\cite{Wang2025MoA}, and MAC-like multi-agent dialogue without cross-debater aggregation~\cite{Chen2025MAC}. Detailed accuracy and safety results for each subset are shown in Table~\ref{tab:performance_comparison}, Table~\ref{tab:performance_comparison_2}, and Fig.~\ref{fig:figure5}. Results on the whole dataset are provided in Supplementary Table~5.

\begin{figure*}[!t]
    \centering
    \includegraphics[width=0.95\textwidth]{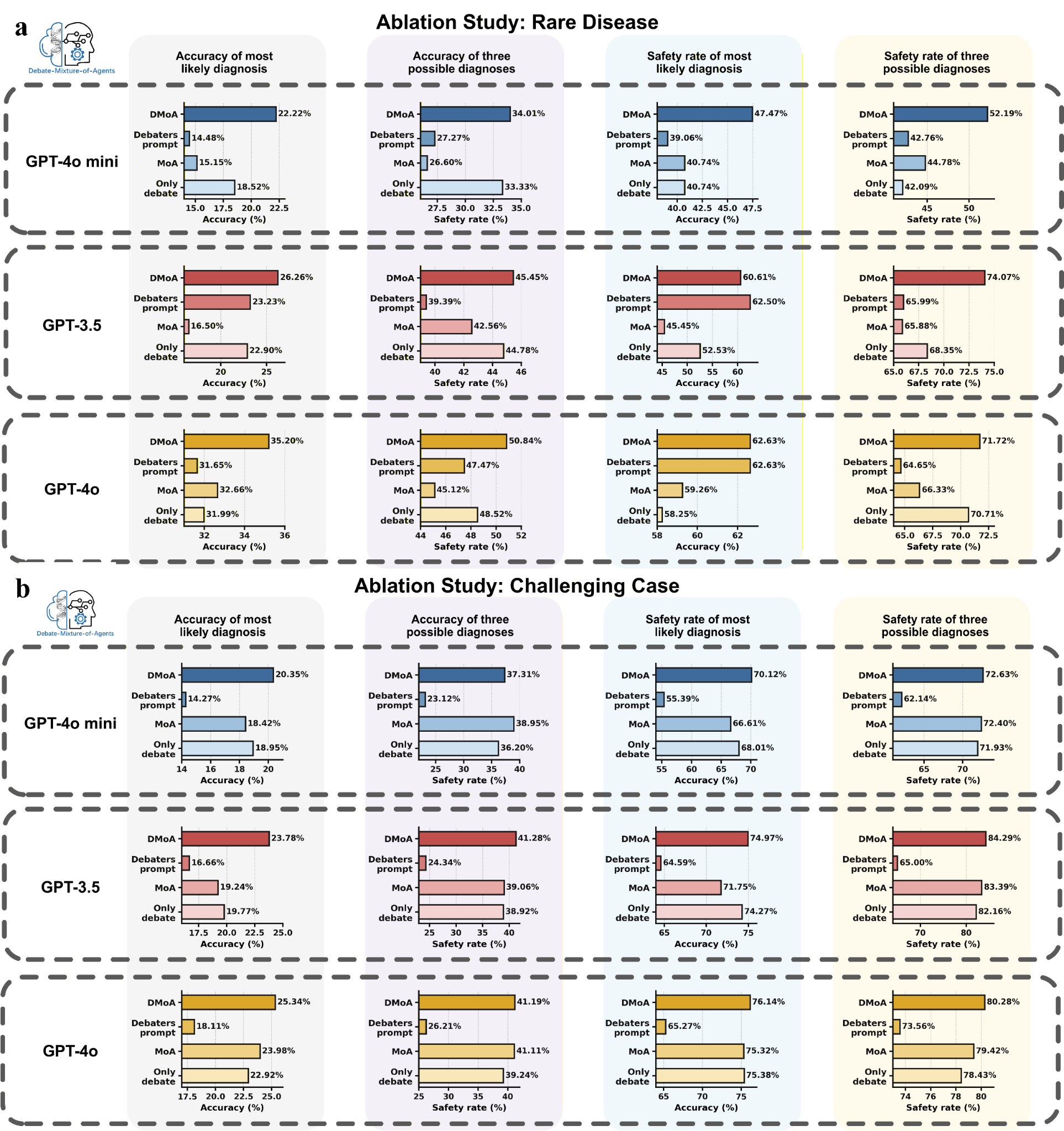}
    \caption[Performance of DMoA and three ablation configurations.]{\textbf{Performance of DMoA and three ablation configurations.}
    \textbf{(a)} Performance in rare disease sub dataset; \textbf{(b)} Performance in challenging cases sub dataset. The bars represent percentages. }
    \label{fig:figure6}
\end{figure*}

Prompt-only emulation did not reproduce the performance of the full DMoA framework. When a single GPT-4o model was prompted to simulate a debate-style reasoning process, it improved over the general prompt in top-1 accuracy (20.10\% vs. 16.58\%) and top-3 accuracy (29.34\% vs. 27.31\%). The corresponding safety rates also increased. However, all four metrics remained lower than those achieved by the 4$\times$2 DMoA framework using the same base model. Similar trends were observed for GPT-3.5 and GPT-4o mini, suggesting that the gain of DMoA cannot be explained by prompt design alone.

Removing the Rebutter reduced diagnostic performance, supporting the contribution of rebuttal-guided revision. When DMoA was downgraded to a simplified MoA structure, the GPT-4o-based framework showed lower top-1 accuracy (25.26\% vs. 26.79\%) and top-1 safety rate (72.95\% vs. 74.15\%) than the full DMoA framework. The same pattern was observed for top-3 diagnosis, with lower accuracy (41.70\% vs. 42.61\%) and safety rate (77.49\% vs. 79.02\%). These results indicate that the Rebutter is not a redundant component, but contributes to diagnostic refinement before aggregation.

Unstructured multi-agent dialogue was also weaker than the full DMoA topology. In the GPT-4o-based setting, the MAC-like multi-turn dialogue framework achieved lower top-1 accuracy (24.26\% vs. 26.79\%) and top-1 safety rate (72.86\% vs. 74.15\%) than DMoA. For top-3 diagnosis, its accuracy (40.61\% vs. 42.61\%) and safety rate (77.29\% vs. 79.02\%) were also lower. Similar trends were observed across other base models. These findings suggest that DMoA benefits not only from multi-agent interaction, but also from its explicit role-constrained topology, in which multiple revised diagnostic hypotheses are preserved and synthesized by the Aggregator rather than being forced into premature consensus.

\subsection{Case-level Analysis of Rebuttal-guided Revision}

To illustrate how the structured topology supported diagnostic refinement, we analyzed a case with a final reference diagnosis of chronic stress-related epiphyseal injury with non-union of the proximal middle phalanx growth plate. The single-model output mainly focused on surface findings, such as finger pain, swelling, and apparent fracture non-union, with limited consideration of repetitive stress injury. In contrast, DMoA progressively refined the diagnosis through rebuttal-guided revision. The Rebutter identified missing evidence, including the absence of trauma, climbing-related repetitive microtrauma, growth plate involvement, CT epiphyseal sclerosis, MRI T2 hyperintensity, and osteochondritis dissecans as an important differential diagnosis. The Presenter then reframed the case as a chronic stress-related epiphyseal or physeal injury, and the Aggregator synthesized the revised hypotheses to produce the final diagnosis. The full intermediate outputs are provided in Supplementary Note~3.

\subsection{Lightweight Model Combination Analysis}
DMoA still showed benefit when lightweight base models were used, but combining a stronger model with lightweight models did not lead to significant improvement. DMoA was evaluated using the same lightweight model, and a mixed setting was also examined in which GPT-3.5-Turbo served as the aggregator while lightweight models were used for the other roles. Detailed results are shown in Supplementary Table 6 and Fig.~\ref{fig:figure7}.

This study investigated DMoA using GLM-4-9B-0414 as the base model in rare disease. Compared to the corresponding single model, DMoA improved the diagnostic performance of GLM-4-9B-0414 in both accuracy and safety rate for most likely diagnosis and three possible diagnoses. The accuracy of the most likely diagnosis increased from 8.08\% to 15.15\%, the safety rate of the most likely diagnosis from 31.99\% to 55.56\%, the accuracy of the three possible diagnoses from 15.49\% to 27.61\%, and the safety rate of the three possible diagnoses from 31.31\% to 42.76\%. All differences were statistically significant (\(p < 0.001\)). Detailed results are shown in Supplementary Table 6.

GPT-3.5 used as the aggregator and lightweight models assigned to the remaining roles was investigated. This mixed setting did not demonstrate a clear advantage over the GPT-3.5 single-model. In rare disease cases, for the most likely diagnosis, both accuracy and safety rate were lower than those of the GPT-3.5-Turbo single model (19.53\% vs. 22.22\%, \(p = 0.9381\); 53.66\% vs. 59.80\%, \(p = 0.9900\)). For the three possible diagnoses, accuracy was numerically higher whereas safety rate was lower (39.66\% vs. 34.68\%, \(p = 0.5450\); 61.49\% vs. 63.97\%, \(p = 0.9732\)). None of these differences reached statistical significance. Detailed results are shown in Supplementary Table 6.

\subsection{Analysis of other Factors Affecting DMoA}
DMoA performance was affected by several configuration factors. Overall, the 4$\times$2 structure achieved the best performance, larger token budgets improved diagnostic performance, and information partitioning did not provide additional benefit. Detailed results are shown in Supplementary Tables~7--9 and Fig.~\ref{fig:figure7}.

\begin{figure*}[!t]
    \centering
    \includegraphics[width=0.85\textwidth]{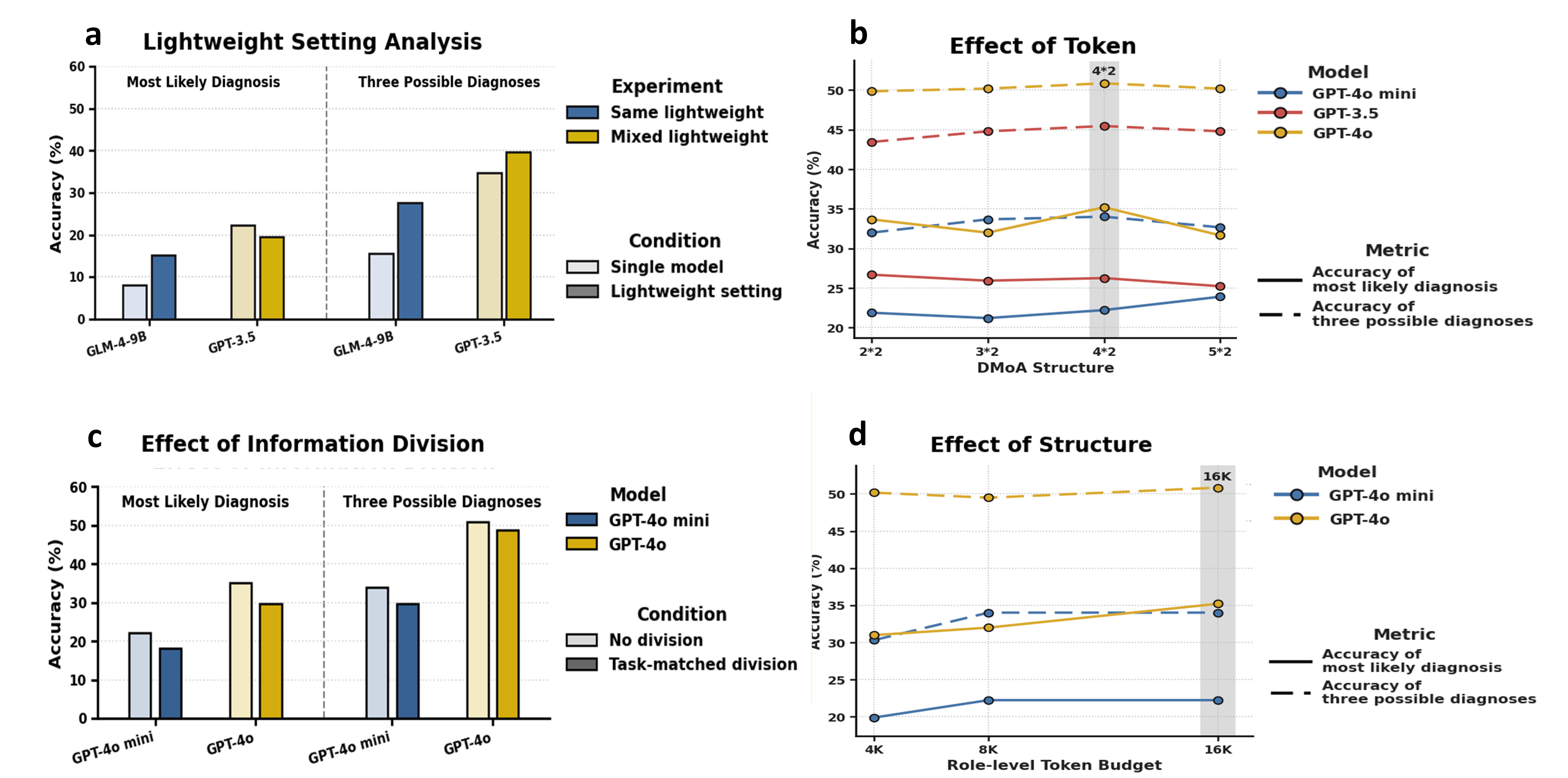}
    \caption[Effects of configuration factors and lightweight settings on DMoA.]{\textbf{Effects of configuration factors and lightweight settings on DMoA.}
    \textbf{(a)} Lightweight-model setting; \textbf{(b)} effect of token budget; \textbf{(c)} effect of information partitioning; \textbf{(d)} effect of DMoA structure. The bars and lines represent percentages.}
    \label{fig:figure7}
\end{figure*}

Using GPT-4o as the base model, the 4$\times$2 structure achieved the highest overall performance among the tested structures. For most likely diagnosis, accuracy reached 35.20\% and safety rate reached 62.63\%, outperforming the 2$\times$2, 3$\times$2, and 5$\times$2 settings. For three possible diagnoses, the 4$\times$2 structure also achieved the highest accuracy and safety rate, at 50.84\% and 71.72\%, respectively. Reducing the token budget from 16,000 to 4,000 decreased top-1 accuracy from 35.20\% to 30.98\% and most likely diagnosis safety rate from 62.63\% to 59.26\%, indicating that sufficient token capacity was important for multi-agent diagnostic reasoning. By contrast, information partitioning did not improve performance. Compared with the non-partitioned setting, the partitioned 3$\times$2 setting showed lower most likely diagnosis accuracy (29.63\% vs. 35.20\%) and similar safety rate (61.28\% vs. 62.63\%). Similar trends were observed for three possible diagnosis. Detailed results are provided in Supplementary Tables~7--9.

Adding a reward model did not consistently improve diagnostic performance. In the GPT-4o-based 4$\times$2 setting, the reward-model variant showed lower most likely diagnosis accuracy than the standard DMoA framework (31.65\% vs. 35.20\%) and similar three possible diagnoses accuracy (50.17\% vs. 50.84\%). Although most likely diagnosis safety rate increased from 62.63\% to 69.36\%, three possible diagnoses safety rate remained essentially unchanged (71.62\% vs. 71.72\%). These results suggest that the tested reward model did not provide a stable overall benefit. Detailed results are shown in Supplementary Table~10.

\subsection{Supplementary Experiments on Clinical Knowledge Q\&A}
DMoA showed little or no advantage over single models in basic clinical knowledge question answering. In the clinical knowledge Q\&A evaluation, both the DMoA framework and the single models achieved more than 70\% accuracy across all four types of questions: 1) Applying Foundational Science Concepts, 2) Practice-based Learning, 3) Professionalism, Legal and Ethical, and 4) Systems-based Practice and Patient Safety. Comparing DMoA with the corresponding single model, GPT-4o achieved total accuracies of 89.03\% vs. 88.00\% (\(p=0.1092\)), GPT-3.5 achieved 81.00\% vs. 79.90\% (\(p=0.1929\)), and GPT-4o mini achieved 74.10\% vs. 75.90\% (\(p=0.9323\)). No significant improvement was observed. Detailed results are shown in Table~\ref{tab:medqa_dimension_accuracy} and Supplementary Table 11.

\begin{table}[!t]
\centering
\caption{Accuracy of single-model and DMoA in clinical Q\&A.}
\label{tab:medqa_dimension_accuracy}
\scriptsize
\renewcommand{\arraystretch}{0.85}
\setlength{\tabcolsep}{1.5pt}
\resizebox{\columnwidth}{!}{
\begin{tabular}{llccccc}
\toprule
\textbf{Model} & \textbf{Structure} 
& \makecell[c]{\textbf{Found.}\\\textbf{Science}}
& \makecell[c]{\textbf{Practice}\\\textbf{Learning}}
& \makecell[c]{\textbf{Professional}\\\textbf{Ethics}}
& \makecell[c]{\textbf{Patient}\\\textbf{Safety}}
& \textbf{Total} \\
\midrule
\multicolumn{7}{c}{\textbf{Single model}} \\
\midrule
GPT-4o mini & NA          & 76.23\% & 74.48\% & 76.87\% & 73.97\% & 75.90\% \\
GPT-3.5     & NA          & 78.80\% & 81.82\% & 80.44\% & 78.80\% & 79.90\% \\
GPT-4o      & NA          & 87.53\% & 87.88\% & 89.38\% & 84.93\% & 88.00\% \\
\midrule
\multicolumn{7}{c}{\textbf{DMoA}} \\
\midrule
GPT-4o mini & $4\times2$ & 73.57\% & 73.84\% & 75.08\% & 73.52\% & 74.10\% \\
GPT-3.5     & $4\times2$ & 80.05\% & 81.82\% & 81.90\% & 82.19\% & 81.00\% \\
GPT-4o      & $4\times2$ & 88.28\% & 89.95\% & 90.01\% & 86.30\% & 89.03\% \\
\bottomrule
\end{tabular}
}
\end{table}

\subsection{Error Analysis}
Diagnostic accuracy errors were grouped into four levels based on their distance from the correct diagnosis. Percentages in the table represent the proportion of all cases. Detailed results are shown in Supplementary Table 12. Each rank of error example are shown in Supplementary Note 2.

\subsection{Reliability Analysis}
The reliability of the single models and DMoA evaluations was assessed across three repeated test rounds using Fleiss' kappa. Results demonstrated only small differences between the testing rounds. In the evaluation of 76 out of 96 single-model and DMoA settings, substantial agreement was observed (0.61--0.80). Detailed results are shown in Supplementary Table 13.

\subsection{Cost Analysis}
The cost analysis of the DMoA framework indicates that for the 4$\times$2 DMoA framework, the average cost per case for the three possible diagnoses was \$0.0089 USD in the rare disease dataset and \$0.0094 USD in the challenging case dataset when using GPT-4o mini as the base model, while it was \$0.1377 USD and \$0.1307 USD per case, respectively, when using GPT-4o as the base model. Detailed results are shown in Supplementary Table 14.

\section{Discussion}
This study developed DMoA, a structured multi-agent framework that reorganizes complex clinical diagnosis into a role-specific and iterative reasoning workflow, and showed that this design improved both diagnostic accuracy and diagnostic safety beyond single models and less structured collaborative settings.

Current LLMs remain inadequate for the demands of real-world complex medical diagnosis. LLMs have shown strong capability in medical question-solving.\cite{singhal2023llmclinicalknowledge} Singhal et al. noted that GPT-4-base had previously achieved an accuracy of 86.1\% on the MedQA dataset.\cite{singhal2025expertqa} However, previous work evaluating standalone models in complex diagnosis showed a clear gap between medical knowledge performance and real diagnostic ability. In rare disease diagnosis, GPT-3.5 and GPT-4 achieved accuracies of only 23\% and 40\%, respectively.\cite{Mehnen2023medRxiv}

One possible reason for this gap lies in the response pattern of LLMs. Current LLMs are still mainly built to give a one-shot response to user input, so they still largely follow a question-answering pattern. \cite{Mondorf2024BeyondAccuracy,Saparov2023GreedyReasoners} This form of reasoning does not match the structured workflow of real clinical diagnosis. Clinical diagnosis is a structured process that requires hypothesis generation, evidence checking, conflict resolution, and revision of conclusions.\cite{Taberna2020FrontOncol,Tu2025Nature} By contrast, a single-model output compresses all of these steps into one response. As a result, even if the model has substantial medical knowledge, it may still fail to perform stable diagnostic reasoning in complex diagnostic cases.\cite{Hager2024NatMed,McDuff2025Nature}

DMoA introduces a more structured organization of diagnostic reasoning to overcome the limitations of the one-shot response pattern. By dividing diagnosis into four role-specific stages: generation, challenge, revision, and aggregation, it distributes key reasoning functions across the debater-aggregator framework rather than relying on a single model to complete complex diagnosis in one pass. This iterative structure may be particularly useful in highly uncertain cases, where diagnosis depends on bringing diverse clinical information together into a coherent explanation. The case-level analysis further supports this mechanism by showing that rebuttal-guided revision can redirect the diagnostic process from surface-level symptom matching toward mechanistic integration of clinical and imaging evidence. Each agent outputs of this example are shown in Supplementary Note 3.

Some recent advances in generative artificial intelligence have already contributed to improving the diagnostic ability of LLMs. Previous studies have shown that prompt design can improve LLM diagnostic performance to some extent. Few-shot prompting, by providing a small number of examples before the target task, has been shown to improve model performance on real clinical cases.\cite{singhal2023llmclinicalknowledge,savage2024diagnosticprompts} Diagnostic-Reasoning Chain-of-Thought (DR-CoT) enhances the organization of the relationship between patient presentation and candidate diagnoses by guiding the model through stepwise clinical reasoning.\cite{savage2024diagnosticprompts,Wu2025MedReason} Similarly, multi-agent discussion can increase the number of viewpoints and promote exchange, which may lead to more diverse information and outputs.\cite{Tang2024MedAgents,Kim2024MDAgents} One previous study also improved LLM diagnostic performance through discussion among multiple doctor agents.\cite{Chen2025MAC}

Further analyses were conducted to examine whether the improved performance of DMoA arose simply from prompt optimization or from increasing the number of agents. Our ablation results showed its improvement depended on specific structured modules. When a single model was prompted to imitate DMoA-style instructions, its performance still remained below that of the full DMoA framework. A previous study also suggested that the effectiveness of multi-agent systems does not arise from prompts alone, but also depends on interaction topology and workflow structure.\cite{Zhou2026MultiAgentDesign} This study also confirmed that multi-round dialogue among multiple agents can improve performance, which is consistent with previous work,\cite{Tang2024MedAgents,Chen2025MAC} but the gain was still smaller than that achieved by structured DMoA. This suggests that explicit internal critique can improve the quality of intermediate diagnostic reasoning and reduce overlooked errors or weakly supported explanations.\cite{Zhao2025NAACL,Irving2018Debate}

Reinforcement learning from human feedback (RLHF) has been used to align LLMs with domain knowledge and user expectations. One study showed that, in general domains, RLHF can improve model responses, making them better aligned with human preferences and instructions.\cite{Ouyang2022InstructGPT} However, its effectiveness in real clinical applications has not yet been fully validated. In this study, a reward model was used to score the Presenter’s responses, but no significant improvement in accuracy or safety rate was observed. This does not mean that the approach is useless. Previous work has suggested several possible explanations. General reward models may not be well suited to medical tasks,\cite{ding2025medrewardbench} and a very small scalar reward signal may provide too little information for the model to use effectively. A previous study also suggested that coarse feedback based only on the final outcome is less effective than richer feedback provided on intermediate reasoning steps.\cite{Lightman2024StepByStep} Although no clear gain was observed in the present study, reward-based alignment may still have potential in medical applications, and improved alignment strategies remain a reasonable direction for future research.\cite{Dong2024RLHFWorkflow}

The structured design of DMoA allows its advantages to become more apparent in complex tasks. In basic clinical question answering, DMoA brought only marginal or no improvement over the corresponding single model. These results indicate that the amplification ability of DMoA for medical factual knowledge is actually quite limited. Previous work has likewise suggested that single models already perform strongly on benchmark-style medical knowledge question answering, whereas the benefit of multi-agent collaboration becomes more evident as task complexity increases. \cite{singhal2025expertqa,Chen2025MAC}

DMoA occupies a distinct position in the field of medical LLMs. LLM-based multi-agent frameworks have already shown encouraging results in medical question answering, multimodal analysis, and medical diagnosis. A series of medical multi-agent frameworks has emerged for different tasks. MDAgents is a multi-expert agent framework designed to improve zero-shot medical question answering through inter-agent communication and consensus building.\cite{Kim2024MDAgents} In MAC, multiple doctor agents engage in multi-round discussion until a supervisor agent determines that consensus has been reached.\cite{Chen2025MAC} Although most existing multi-agent frameworks rely mainly on communication to reach consensus, the present framework establishes a more structured division of labor and focuses on complex clinical diagnostic tasks. By contrast, the collaborative diagnostic reasoning process in DMoA is not limited to simple interaction, but is organized more explicitly. DMoA also does not force all agents to agree. It allows different, and even conflicting, candidate results to be retained and then filtered through aggregation by the Aggregator. The structured design adopted in this study therefore differs from consensus-only interaction. Although the goals and characteristics of these agent systems are not identical, their collective potential in medicine appears substantial.\cite{Zhou2025Scoping} Further work is needed to explore and evaluate their capabilities more fully in medical settings.\cite{Busch2025CommMed,Zhou2025Scoping}

Our investigation into the DMoA framework revealed several other factors that influenced its performance. The choice of base model remained an important determinant of overall performance, suggesting that the upper bound of a collaborative system is still constrained by the capability of its component models. The structural complexity of DMoA also affected the overall framework performance, with the 4 $\times$ 2 configuration achieving the best results. Increasing structural complexity did not always lead to further gains. Previous work has suggested that multi-agent systems may have an optimal scale, beyond which marginal returns decline.\cite{yang2026agentscaling} Heterogeneous-model experiments further showed that the way models are combined also matters. Models with more similar capability may support more effective collaboration, whereas large performance gaps between roles do not necessarily produce better results. Previous work has offered a similar explanation: the effectiveness of heterogeneous multi-agent collaboration depends on whether weaker agents can understand and execute information effectively.\cite{Wang2026GuidedCollaboration} Lightweight-model experiments revealed a practically meaningful deployment strategy. Replacing base model with lightweight models reduced absolute performance, but the resulting system still outperformed its corresponding single-model baseline. The analyses of token budget and information allocation indicate that the performance of the DMoA framework may be adversely affected by an excessively low token limit.

Many recent studies have proposed evaluation methods for LLM diagnostic outputs.\cite{Bedi2025JAMA,Takita2025Meta,Johri2025NatMed} This study adopted a highly strict criterion: a prediction was considered accurate only when it exactly matched the reference diagnosis for accuracy evaluation, which partly explains why the reported accuracy values were relatively low. Clinical diagnosis should not be judged by accuracy alone, because diagnostic safety is also critical.\cite{Omar2025CommMed,Hager2024NatMed} Previous studies have already begun to incorporate safety into evaluation.\cite{Bedi2025JAMA,Johri2025NatMed} One prior study proposed a framework for assessing clinical safety and hallucination in LLM-generated medical summaries.\cite{Asgari2025ClinicalSafetyHallucinationSummarisation} In the present study, a structured scoring framework was used to evaluate both diagnostic accuracy and five safety dimensions, providing a more comprehensive way to quantify LLM-generated clinical natural-language outputs. 

McDuff et al. reported that unassisted clinicians achieved a top-10 accuracy of 33.6\% on challenging published case reports \cite{McDuff2025Nature}. However, direct numerical comparison with our results is not appropriate: our top-3 metric is evaluated under a stricter recall constraint than top-10, and the two studies differ in case source, case selection criteria, and evaluation protocol. These differences preclude any inference about relative performance between DMoA and human clinicians.

This study still has several limitations. Although a relatively large dataset of difficult cases was included, the number of rare disease samples still remains limited relative to the large number of rare diseases overall.\cite{Haendel2020HowManyRareDiseases,Richter2015ValueHealth} All cases were derived from published reports rather than prospective clinical workflows. The test setting therefore may not always fully reflect real diagnostic practice. The framework also still depends on the capability of the base model. When lightweight models were used as the foundation, performance remained better than the corresponding single-model baseline, but the absolute level was still constrained.

The role of multi-model collaboration in medicine can vary across applications. AgentClinic focuses on multimodal medical diagnosis,\cite{Schmidgall2024AgentClinic} whereas MEDAGENTS is designed for medical question answering.\cite{Tang2024MedAgents} Despite the above limitations, DMoA may still serve as a structured framework for high-difficulty, high-uncertainty cases. In practical use, it could be further integrated with agents for upstream data-processing tasks, such as case-processing agents and gene-analysis agents, to improve specialization and real-world performance. Hospitals may also deploy locally hosted small-model versions as lightweight implementations for data safety that reduce cost while still achieving performance suitable for routine use. It should also be noted that, because interpretability still has clear boundaries and current evaluation schemes remain incomplete, DMoA cannot fully replace clinicians and should instead be used as a second-opinion tool.\cite{Bedi2025JAMA,Johri2025NatMed}

In conclusion, this study shows that the DMoA framework substantially improves the diagnostic ability of LLMs in complex clinical tasks. In real diagnostic settings, standalone models still have difficulty coordinating evidence and sustaining deep reasoning.\cite{Hager2024NatMed,McDuff2025Nature} DMoA addresses this gap by organizing diagnostic reasoning into four explicit stages—generation, critique, revision, and aggregation—which significantly improves diagnostic performance and helps compensate for the reasoning limits of one-shot single-model outputs. These results highlight the value of structured multi-agent systems in medical diagnosis and support further investigation into their design.\cite{Chen2025MAC,Tang2024MedAgents,Kim2024MDAgents}

\section*{Data and Code Availability}
The rare disease cases and challenging case datasets are from The ClinDiag code repository which is released under the Apache-2.0 license, whereas the dataset should be used according to the license and data-use terms specified by the original authors. The clinical Q\&A dataset is filtered from MedQA dataset and is publicly available at \url{https://github.com/jind11/MedQA} under the MIT license.

The Dataset is openly available at the following link for non-commercial purpose: \url{https://github.com/Ivesxia/Debate-Mixture-of-Agent}

Code is openly available at the following link for non-commercial purpose: \url{https://github.com/Ivesxia/Debate-Mixture-of-Agent}

\section*{Conflict of Interest Statement}
The authors declare no conflicts of interest.

\section*{References}

% \bibliographystyle{IEEEtran}
% \bibliography{references}

% Generated by IEEEtran.bst, version: 1.14 (2015/08/26)
\begin{thebibliography}{10}
\providecommand{\url}[1]{#1}
\csname url@samestyle\endcsname
\providecommand{\newblock}{\relax}
\providecommand{\bibinfo}[2]{#2}
\providecommand{\BIBentrySTDinterwordspacing}{\spaceskip=0pt\relax}
\providecommand{\BIBentryALTinterwordstretchfactor}{4}
\providecommand{\BIBentryALTinterwordspacing}{\spaceskip=\fontdimen2\font plus
\BIBentryALTinterwordstretchfactor\fontdimen3\font minus \fontdimen4\font\relax}
\providecommand{\BIBforeignlanguage}[2]{{%
\expandafter\ifx\csname l@#1\endcsname\relax
\typeout{** WARNING: IEEEtran.bst: No hyphenation pattern has been}%
\typeout{** loaded for the language `#1'. Using the pattern for}%
\typeout{** the default language instead.}%
\else
\language=\csname l@#1\endcsname
\fi
#2}}
\providecommand{\BIBdecl}{\relax}
\BIBdecl

\bibitem{VanVeen2024NatMed}
\BIBentryALTinterwordspacing
D.~Van~Veen, C.~Van~Uden, L.~Blankemeier, J.-B. Delbrouck, A.~Aali, C.~Bluethgen, A.~Pareek, M.~Polacin, E.~P. Reis, A.~Seehofnerov{\'a}, N.~Rohatgi, P.~Hosamani, W.~Collins, N.~Ahuja, C.~P. Langlotz, J.~Hom, S.~Gatidis, J.~Pauly, and A.~S. Chaudhari, ``Adapted large language models can outperform medical experts in clinical text summarization,'' \emph{Nature Medicine}, vol.~30, no.~4, pp. 1134--1142, 2024. [Online]. Available: \url{https://doi.org/10.1038/s41591-024-02855-5}
\BIBentrySTDinterwordspacing

\bibitem{Busch2025CommMed}
\BIBentryALTinterwordspacing
F.~Busch, L.~Hoffmann, C.~Rueger, E.~H. van Dijk, R.~Kader, E.~Ortiz-Prado, M.~R. Makowski, L.~Saba, M.~Hadamitzky, J.~N. Kather, D.~Truhn, R.~Cuocolo, L.~C. Adams, and K.~K. Bressem, ``Current applications and challenges in large language models for patient care: a systematic review,'' \emph{Communications Medicine}, vol.~5, no.~1, p.~26, 2025. [Online]. Available: \url{https://doi.org/10.1038/s43856-024-00717-2}
\BIBentrySTDinterwordspacing

\bibitem{Zhou2025Scoping}
\BIBentryALTinterwordspacing
S.~Zhou, Z.~Xu, M.~Zhang, C.~Xu, Y.~Guo, Z.~Zhan, Y.~Fang, S.~Ding, J.~Wang, K.~Xu, L.~Xia, J.~Yeung, D.~Zha, D.~Cai, G.~B. Melton, M.~Lin, and R.~Zhang, ``Large language models for disease diagnosis: a scoping review,'' \emph{npj Artificial Intelligence}, vol.~1, no.~1, p.~9, 2025. [Online]. Available: \url{https://doi.org/10.1038/s44387-025-00011-z}
\BIBentrySTDinterwordspacing

\bibitem{Bedi2025JAMA}
S.~Bedi, Y.~Liu, L.~Orr-Ewing \emph{et~al.}, ``Testing and evaluation of health care applications of large language models: A systematic review,'' \emph{JAMA}, vol. 333, no.~4, pp. 319--328, 2025.

\bibitem{Tu2025Nature}
T.~Tu, M.~Schaekermann, A.~Palepu \emph{et~al.}, ``Towards conversational diagnostic artificial intelligence,'' \emph{Nature}, vol. 642, no. 8067, pp. 442--450, 2025.

\bibitem{McDuff2025Nature}
D.~McDuff, M.~Schaekermann, T.~Tu \emph{et~al.}, ``Towards accurate differential diagnosis with large language models,'' \emph{Nature}, vol. 642, no. 8067, pp. 451--457, 2025.

\bibitem{Faye2024EJHG}
\BIBentryALTinterwordspacing
F.~Faye, C.~Crocione, R.~Anido~de Pe{\~n}a, S.~Bellagambi, L.~Escati~Pe{\~n}aloza, A.~Hunter, L.~Jensen, C.~Oosterwijk, E.~Schoeters, D.~de~Vicente, L.~Faivre, M.~Wilbur, Y.~Le~Cam, and J.~Dubief, ``Time to diagnosis and determinants of diagnostic delays of people living with a rare disease: results of a rare barometer retrospective patient survey,'' \emph{European Journal of Human Genetics}, vol.~32, no.~9, pp. 1116--1126, 2024. [Online]. Available: \url{https://doi.org/10.1038/s41431-024-01604-z}
\BIBentrySTDinterwordspacing

\bibitem{Phillips2024OJRD}
\BIBentryALTinterwordspacing
C.~Phillips, A.~Parkinson, T.~Namsrai, A.~Chalmers, C.~Dews, D.~Gregory, E.~Kelly, C.~Lowe, and J.~Desborough, ``Time to diagnosis for a rare disease: managing medical uncertainty. a qualitative study,'' \emph{Orphanet Journal of Rare Diseases}, vol.~19, no.~1, p. 297, 2024. [Online]. Available: \url{https://doi.org/10.1186/s13023-024-03319-2}
\BIBentrySTDinterwordspacing

\bibitem{Hager2024NatMed}
\BIBentryALTinterwordspacing
P.~Hager, F.~Jungmann, R.~Holland, K.~Bhagat, I.~Hubrecht, M.~Knauer, J.~Vielhauer, M.~Makowski, R.~Braren, G.~Kaissis, and D.~Rueckert, ``Evaluation and mitigation of the limitations of large language models in clinical decision-making,'' \emph{Nature Medicine}, vol.~30, no.~9, pp. 2613--2622, 2024. [Online]. Available: \url{https://doi.org/10.1038/s41591-024-03097-1}
\BIBentrySTDinterwordspacing

\bibitem{Tang2024MedAgents}
\BIBentryALTinterwordspacing
X.~Tang, A.~Zou, Z.~Zhang, Z.~Li, Y.~Zhao, X.~Zhang, A.~Cohan, and M.~Gerstein, ``Medagents: Large language models as collaborators for zero-shot medical reasoning,'' in \emph{Findings of the Association for Computational Linguistics: ACL 2024}.\hskip 1em plus 0.5em minus 0.4em\relax Bangkok, Thailand: Association for Computational Linguistics, Aug. 2024, pp. 599--621. [Online]. Available: \url{https://aclanthology.org/2024.findings-acl.33/}
\BIBentrySTDinterwordspacing

\bibitem{Chen2025MAC}
\BIBentryALTinterwordspacing
X.~Chen, H.~Yi, M.~You, W.~Liu, L.~Wang, H.~Li, X.~Zhang, Y.~Guo, L.~Fan, G.~Chen, Q.~Lao, W.~Fu, K.~Li, and J.~Li, ``Enhancing diagnostic capability with multi-agents conversational large language models,'' \emph{npj Digital Medicine}, vol.~8, no.~1, p. 159, 2025. [Online]. Available: \url{https://doi.org/10.1038/s41746-025-01550-0}
\BIBentrySTDinterwordspacing

\bibitem{Shazeer2017MoE}
\BIBentryALTinterwordspacing
N.~Shazeer, A.~Mirhoseini, K.~Maziarz, A.~Davis, Q.~V. Le, G.~E. Hinton, and J.~Dean, ``Outrageously large neural networks: The sparsely-gated mixture-of-experts layer,'' in \emph{International Conference on Learning Representations (ICLR)}, 2017. [Online]. Available: \url{https://arxiv.org/abs/1701.06538}
\BIBentrySTDinterwordspacing

\bibitem{Wang2025MoA}
\BIBentryALTinterwordspacing
J.~Wang, J.~Wang, B.~Athiwaratkun, C.~Zhang, and J.~Zou, ``Mixture-of-agents enhances large language model capabilities,'' in \emph{International Conference on Learning Representations (ICLR)}, 2025. [Online]. Available: \url{https://openreview.net/forum?id=h0ZfDIrj7T}
\BIBentrySTDinterwordspacing

\bibitem{Li2025GroundingLLMClinicalDiagnostics}
\BIBentryALTinterwordspacing
J.~Li, X.~Chen, H.~Zhou, H.~Yi, M.~You, W.~Liu, L.~Wang, H.~Li, X.~Zhang, Y.~Guo, L.~Fan, Q.~Lao, W.~Fu, and K.~Li, ``Grounding large language model in clinical diagnostics,'' \emph{Research Square}, Apr. 2025, posted Date: 2025-04-15; preprint version v1. [Online]. Available: \url{https://doi.org/10.21203/rs.3.rs-6286021/v1}
\BIBentrySTDinterwordspacing

\bibitem{Jin2020MedQA}
\BIBentryALTinterwordspacing
D.~Jin, E.~Pan, N.~Oufattole, W.-H. Weng, H.~Fang, and P.~Szolovits, ``What disease does this patient have? a large-scale open domain question answering dataset from medical exams,'' 2020. [Online]. Available: \url{https://arxiv.org/abs/2009.13081}
\BIBentrySTDinterwordspacing

\bibitem{USMLEPhysicianTasksCompetencies}
\BIBentryALTinterwordspacing
{USMLE}, ``{USMLE} physician tasks/competencies,'' PDF, 2021. [Online]. Available: \url{https://www.usmle.org/sites/default/files/2021-08/USMLE_Physician_Tasks_Competencies.pdf}
\BIBentrySTDinterwordspacing

\bibitem{maiga-etal-2025-error}
A.~Maiga, A.~Shah, and E.~Yilmaz, ``Error detection in medical note through multi agent debate,'' in \emph{Proceedings of the 24th Workshop on Biomedical Language Processing}, Aug. 2025, pp. 124--135.

\bibitem{Likert1932}
R.~Likert, ``A technique for the measurement of attitudes,'' \emph{Archives of Psychology}, vol. 140, pp. 1--55, 1932.

\bibitem{Bond2012JGIM}
\BIBentryALTinterwordspacing
W.~F. Bond, L.~M. Schwartz, K.~R. Weaver, D.~Levick, M.~Giuliano, and M.~L. Graber, ``Differential diagnosis generators: an evaluation of currently available computer programs,'' \emph{Journal of General Internal Medicine}, vol.~27, no.~2, pp. 213--219, 2012. [Online]. Available: \url{https://doi.org/10.1007/s11606-011-1804-8}
\BIBentrySTDinterwordspacing

\bibitem{Asgari2025ClinicalSafetyHallucinationSummarisation}
\BIBentryALTinterwordspacing
E.~Asgari, N.~Monta{\~n}a-Brown, M.~Dubois, S.~Khalil, J.~Balloch, J.~Au~Yeung, and D.~Pimenta, ``A framework to assess clinical safety and hallucination rates of {LLMs} for medical text summarisation,'' \emph{npj Digital Medicine}, vol.~8, no.~1, p. 274, 2025. [Online]. Available: \url{https://www.nature.com/articles/s41746-025-01670-7}
\BIBentrySTDinterwordspacing

\bibitem{Wu2025MedReason}
J.~Wu, W.~Deng, X.~Li, S.~Liu, T.~Mi, Y.~Peng, Z.~Xu, Y.~Liu, H.~Cho, C.-I. Choi, Y.~Cao, H.~Ren, X.~Li, X.~Li, and Y.~Zhou, ``Medreason: Eliciting factual medical reasoning steps in llms via knowledge graphs,'' \emph{arXiv preprint arXiv:2504.00993}, Apr. 2025.

\bibitem{Fleiss1971Kappa}
\BIBentryALTinterwordspacing
J.~L. Fleiss, ``Measuring nominal scale agreement among many raters,'' \emph{Psychological Bulletin}, vol.~76, no.~5, pp. 378--382, 1971. [Online]. Available: \url{https://doi.org/10.1037/h0031619}
\BIBentrySTDinterwordspacing

\bibitem{Landis1977Kappa}
\BIBentryALTinterwordspacing
J.~R. Landis and G.~G. Koch, ``The measurement of observer agreement for categorical data,'' \emph{Biometrics}, vol.~33, no.~1, pp. 159--174, 1977. [Online]. Available: \url{https://doi.org/10.2307/2529310}
\BIBentrySTDinterwordspacing

\bibitem{singhal2023llmclinicalknowledge}
K.~Singhal, S.~Azizi, T.~Tu, S.~S. Mahdavi, J.~Wei, H.~W. Chung, N.~Scales, A.~Tanwani, H.~Cole-Lewis, S.~Pfohl \emph{et~al.}, ``Large language models encode clinical knowledge,'' \emph{Nature}, vol. 620, no. 7972, pp. 172--180, 2023.

\bibitem{singhal2025expertqa}
K.~Singhal, T.~Tu, J.~Gottweis \emph{et~al.}, ``Toward expert-level medical question answering with large language models,'' \emph{Nature Medicine}, vol.~31, no.~3, pp. 943--950, 2025.

\bibitem{Mehnen2023medRxiv}
\BIBentryALTinterwordspacing
L.~Mehnen, S.~Gruarin, M.~Vasileva, and B.~Knapp, ``Chatgpt as a medical doctor? a diagnostic accuracy study on common and rare diseases,'' \emph{medRxiv}, 2023. [Online]. Available: \url{https://doi.org/10.1101/2023.04.20.23288859}
\BIBentrySTDinterwordspacing

\bibitem{Mondorf2024BeyondAccuracy}
\BIBentryALTinterwordspacing
P.~Mondorf and B.~Plank, ``Beyond accuracy: Evaluating the reasoning behavior of large language models -- a survey,'' 2024. [Online]. Available: \url{https://arxiv.org/abs/2404.01869}
\BIBentrySTDinterwordspacing

\bibitem{Saparov2023GreedyReasoners}
\BIBentryALTinterwordspacing
A.~Saparov and H.~He, ``Language models are greedy reasoners: A systematic formal analysis of chain-of-thought,'' in \emph{International Conference on Learning Representations (ICLR)}, 2023. [Online]. Available: \url{https://arxiv.org/abs/2210.01240}
\BIBentrySTDinterwordspacing

\bibitem{Taberna2020FrontOncol}
\BIBentryALTinterwordspacing
M.~Taberna, F.~Gil~Moncayo, E.~Jan{\'e}-Salas, M.~Antonio, L.~Arribas, E.~Vilajosana, E.~Peralvez~Torres, and R.~Mes{\'i}a, ``The multidisciplinary team ({MDT}) approach and quality of care,'' \emph{Frontiers in Oncology}, vol.~10, p.~85, 2020. [Online]. Available: \url{https://doi.org/10.3389/fonc.2020.00085}
\BIBentrySTDinterwordspacing

\bibitem{savage2024diagnosticprompts}
T.~Savage, A.~Nayak, R.~N. Gallo, E.~S. Rangan, and J.~H. Chen, ``Diagnostic reasoning prompts reveal the potential for large language model interpretability in medicine,'' \emph{npj Digital Medicine}, vol.~7, p.~20, 2024.

\bibitem{Kim2024MDAgents}
Y.~Kim, C.~Park, H.~Jeong, Y.~S. Chan, X.~Xu, D.~McDuff, H.~Lee, M.~Ghassemi, C.~Breazeal, and H.~W. Park, ``Mdagents: An adaptive collaboration of llms for medical decision-making,'' in \emph{Advances in Neural Information Processing Systems}, vol.~37, 2024.

\bibitem{Zhou2026MultiAgentDesign}
\BIBentryALTinterwordspacing
H.~Zhou, X.~Wan, R.~Sun, H.~Palangi, S.~Iqbal, I.~Vuli{\'c}, A.~Korhonen, and S.~O. Arik, ``Multi-agent design: Optimizing agents with better prompts and topologies,'' in \emph{International Conference on Learning Representations (ICLR)}, 2026, poster. [Online]. Available: \url{https://openreview.net/forum?id=I05H9RUzHB}
\BIBentrySTDinterwordspacing

\bibitem{Zhao2025NAACL}
\BIBentryALTinterwordspacing
Y.~Zhao, H.~Wang, Y.~Zheng, and X.~Wu, ``A layered debating multi-agent system for similar disease diagnosis,'' in \emph{Proceedings of the 2025 Conference of the Nations of the Americas Chapter of the Association for Computational Linguistics: Human Language Technologies (Volume 2: Short Papers)}.\hskip 1em plus 0.5em minus 0.4em\relax Albuquerque, New Mexico: Association for Computational Linguistics, Apr. 2025, pp. 539--549. [Online]. Available: \url{https://aclanthology.org/2025.naacl-short.46/}
\BIBentrySTDinterwordspacing

\bibitem{Irving2018Debate}
\BIBentryALTinterwordspacing
G.~Irving, P.~Christiano, and D.~Amodei, ``Ai safety via debate,'' 2018. [Online]. Available: \url{https://arxiv.org/abs/1805.00899}
\BIBentrySTDinterwordspacing

\bibitem{Ouyang2022InstructGPT}
L.~Ouyang, J.~Wu, X.~Jiang, D.~Almeida, C.~L. Wainwright, P.~Mishkin, C.~Zhang, S.~Agarwal, K.~Slama, A.~Ray, J.~Schulman, J.~Hilton, F.~Kelton, L.~Miller, M.~Simens, A.~Askell, P.~Welinder, P.~Christiano, J.~Leike, and R.~Lowe, ``Training language models to follow instructions with human feedback,'' in \emph{Advances in Neural Information Processing Systems}, vol.~35, 2022.

\bibitem{ding2025medrewardbench}
M.~Ding, J.~Zhang, W.~Wang, C.-Y. Li, W.-C. Fang, H.-Y. Wu, H.~Zhong, W.~Chen, and L.~Shen, ``Med-rewardbench: Benchmarking reward models and judges for medical multimodal large language models,'' \emph{arXiv preprint arXiv:2508.21430}, 2025.

\bibitem{Lightman2024StepByStep}
\BIBentryALTinterwordspacing
H.~Lightman, V.~Kosaraju, Y.~Burda, H.~Edwards, B.~Baker, T.~Lee, J.~Leike, J.~Schulman, I.~Sutskever, and K.~Cobbe, ``Let's verify step by step,'' in \emph{The Twelfth International Conference on Learning Representations (ICLR)}, 2024. [Online]. Available: \url{https://openreview.net/forum?id=v8L0pN6EOi}
\BIBentrySTDinterwordspacing

\bibitem{Dong2024RLHFWorkflow}
\BIBentryALTinterwordspacing
H.~Dong, W.~Xiong, B.~Pang, H.~Wang, H.~Zhao, Y.~Zhou, N.~Jiang, D.~Sahoo, C.~Xiong, and T.~Zhang, ``Rlhf workflow: From reward modeling to online rlhf,'' 2024. [Online]. Available: \url{https://arxiv.org/abs/2405.07863}
\BIBentrySTDinterwordspacing

\bibitem{yang2026agentscaling}
Y.~Yang, C.~Qu, M.~Wen, L.~Shi, Y.~Wen, W.~Zhang, A.~Wierman, and S.~Gu, ``Understanding agent scaling in llm-based multi-agent systems via diversity,'' \emph{arXiv preprint arXiv:2602.03794}, 2026.

\bibitem{Wang2026GuidedCollaboration}
\BIBentryALTinterwordspacing
L.~Wang, T.~Zhu, L.~Qin, L.~Gao, and W.~Zhou, ``Guided collaboration in heterogeneous {LLM}-based multi-agent systems via entropy-based understanding assessment and experience retrieval,'' \emph{arXiv preprint arXiv:2602.13639}, 2026. [Online]. Available: \url{https://arxiv.org/abs/2602.13639}
\BIBentrySTDinterwordspacing

\bibitem{Takita2025Meta}
H.~Takita, D.~Kabata, S.~L. Walston, H.~Tatekawa, K.~Saito, Y.~Tsujimoto, Y.~Miki, and D.~Ueda, ``A systematic review and meta-analysis of diagnostic performance comparison between generative ai and physicians,'' \emph{npj Digital Medicine}, vol.~8, p. 175, 2025.

\bibitem{Johri2025NatMed}
S.~Johri, J.~Jeong, B.~A. Tran \emph{et~al.}, ``An evaluation framework for clinical use of large language models in patient interaction tasks,'' \emph{Nature Medicine}, vol.~31, pp. 77--86, 2025.

\bibitem{Omar2025CommMed}
\BIBentryALTinterwordspacing
M.~Omar, V.~Sorin, J.~D. Collins, D.~Reich, R.~Freeman, N.~Gavin, A.~Charney, L.~Stump, N.~L. Bragazzi, G.~N. Nadkarni, and E.~Klang, ``Multi-model assurance analysis showing large language models are highly vulnerable to adversarial hallucination attacks during clinical decision support,'' \emph{Communications Medicine}, vol.~5, no.~1, p. 330, 2025. [Online]. Available: \url{https://doi.org/10.1038/s43856-025-01021-3}
\BIBentrySTDinterwordspacing

\bibitem{Haendel2020HowManyRareDiseases}
\BIBentryALTinterwordspacing
M.~Haendel, N.~Vasilevsky, D.~Unni, C.~Bologa, N.~Harris, H.~Rehm, A.~Hamosh, G.~Baynam, T.~Groza, J.~McMurry, H.~Dawkins, A.~Rath, C.~Thaxon, G.~Bocci, M.~P. Joachimiak, S.~K{\"o}hler, P.~N. Robinson, C.~Mungall, and T.~I. Oprea, ``How many rare diseases are there?'' \emph{Nature Reviews Drug Discovery}, vol.~19, no.~2, pp. 77--78, 2020. [Online]. Available: \url{https://doi.org/10.1038/d41573-019-00180-y}
\BIBentrySTDinterwordspacing

\bibitem{Richter2015ValueHealth}
\BIBentryALTinterwordspacing
T.~Richter, S.~Nestler-Parr, R.~Babela, Z.~M. Khan, T.~Tesoro, E.~Molsen, and D.~A. Hughes, ``Rare disease terminology and definitions---a systematic global review: Report of the {ISPOR} rare disease special interest group,'' \emph{Value in Health}, vol.~18, no.~6, pp. 906--914, 2015. [Online]. Available: \url{https://doi.org/10.1016/j.jval.2015.05.008}
\BIBentrySTDinterwordspacing

\bibitem{Schmidgall2024AgentClinic}
\BIBentryALTinterwordspacing
S.~Schmidgall, R.~Ziaei, C.~Harris, E.~Reis, J.~Jopling, and M.~Moor, ``Agentclinic: a multimodal agent benchmark to evaluate ai in simulated clinical environments,'' 2024. [Online]. Available: \url{https://arxiv.org/abs/2405.07960}
\BIBentrySTDinterwordspacing

\end{thebibliography}

\end{document}